*Article*

# Stride Independent Patching for Deep Learning

**Olivier Rukundo**

Department of Electronic and Computer Engineering; University of Limerick; e-mail: olivier.rukundo@ul.ie

**Abstract:** This paper presents semi-automatic stride-independent patching (SSP) as an alternative to automatic stride-dependent patching techniques. SSP uses user or expert input to position pre-defined patches over one or more objects of interest. To evaluate its effectiveness, three patch-based datasets were created using SSP, overlapping patching (Overlap), and non-overlapping patching (Noverlap). DeepLabV3+ models with ResNet50, ResNet18, and MobileNetV2 backbones were trained separately on each dataset. Quantitative evaluations were performed on the respective test splits and a common external test set. SSP generally achieved higher segmentation scores on the test splits and required the shortest model training time across all three backbones. On the external test set, SSP achieved the highest average precision and F1-score across backbones, whereas Noverlap achieved the highest average recall. These preliminary results demonstrate that the potentially greater spatial coverage of Noverlap and Overlap did not generally translate into better segmentation performance and that SSP offers a favorable balance between segmentation performance and model training time.

## 1. Introduction

Image patching techniques provide an alternative approach that makes training deep learning models on large or high-resolution images feasible under tight hardware constraints [1], [2], [26-29]. As of today, there are two traditional image patching techniques that are still popular or widely used to create image patches for deep learning model training purposes, namely: non-overlapping and overlapping techniques. However, all these techniques remain stride-dependent and often produce redundant or irrelevant patches, which may lead to inefficient training and suboptimal accuracy. A stride is the step size by which the small image patch, of a specific or pre-defined size, moves across a large image [5], [6], [30], [31], [50–53]. In this context, a stride decides the location of image patches to extract but does not automatically identify the patch-of-interest - unless guided by additional modern mechanisms [4]. Modern mechanisms can help identify patches of interest (POIs), but at an additional computational cost and with a risk of failing to fully include objects of interest within those patches. To the best of the author's knowledge: when stride < patch size, overlapping patches are created resulting in excessive number of patches that may introduce redundancy and increasing computational cost. When, stride = patch size: Non-overlapping patches are created and the resulting patching may possibly cut through objects arbitrarily, often ignoring object of interest boundaries, which can lead to fragmented and incomplete segmentation [3]. When stride > patch size: Gaps between patches are created and, in this context, this stride setting cannot be an option.

In efforts to preserve objects of interest or their groups' boundaries, avoid redundant patches, remove or mitigate the risk of losing objects and finally better leverage limited memory hardware without sacrificing the accuracy and efficiency, an interactive tool is introduced for semi-automatic stride-independent patching (SSP). In brief, the tool leverages the expert user's capability to locate and select patch-of-interest containing objects of interest. Here, objects of interests are adenoviruses [7], [32], [33]. The SSP allows users to inspect and adjust patch placement to fully include objects of interest, reducing the risk of object omission or fragmentation during patch selection. Unlike the adaptive patching in [4], the introduced SSP tool does not rely on down-scaling operation at any stage, instead it directly leverages the expert user's capability to fit the patch-of-interest around objects of interest that completely fit inside the pre-defined image patch size.

## 2. Materials and Methods

### *2.1. SSP tool description*

The mask-guided semi-automatic stride-independent patching (SSP) tool was implemented in Python using OpenCV, Matplotlib, and Tkinter in Anaconda's Spyder environment [54]. The tool combines interactive patch selection and image–mask patch extraction in a single script, ssp26.py [54]. Figure 1 shows the main parts of the SSP tool. Users position fixed-size square patches around objects of interest visible in an existing segmentation mask. Identical crop or patch coordinates are then applied to the mask and its corresponding image to preserve spatial alignment.

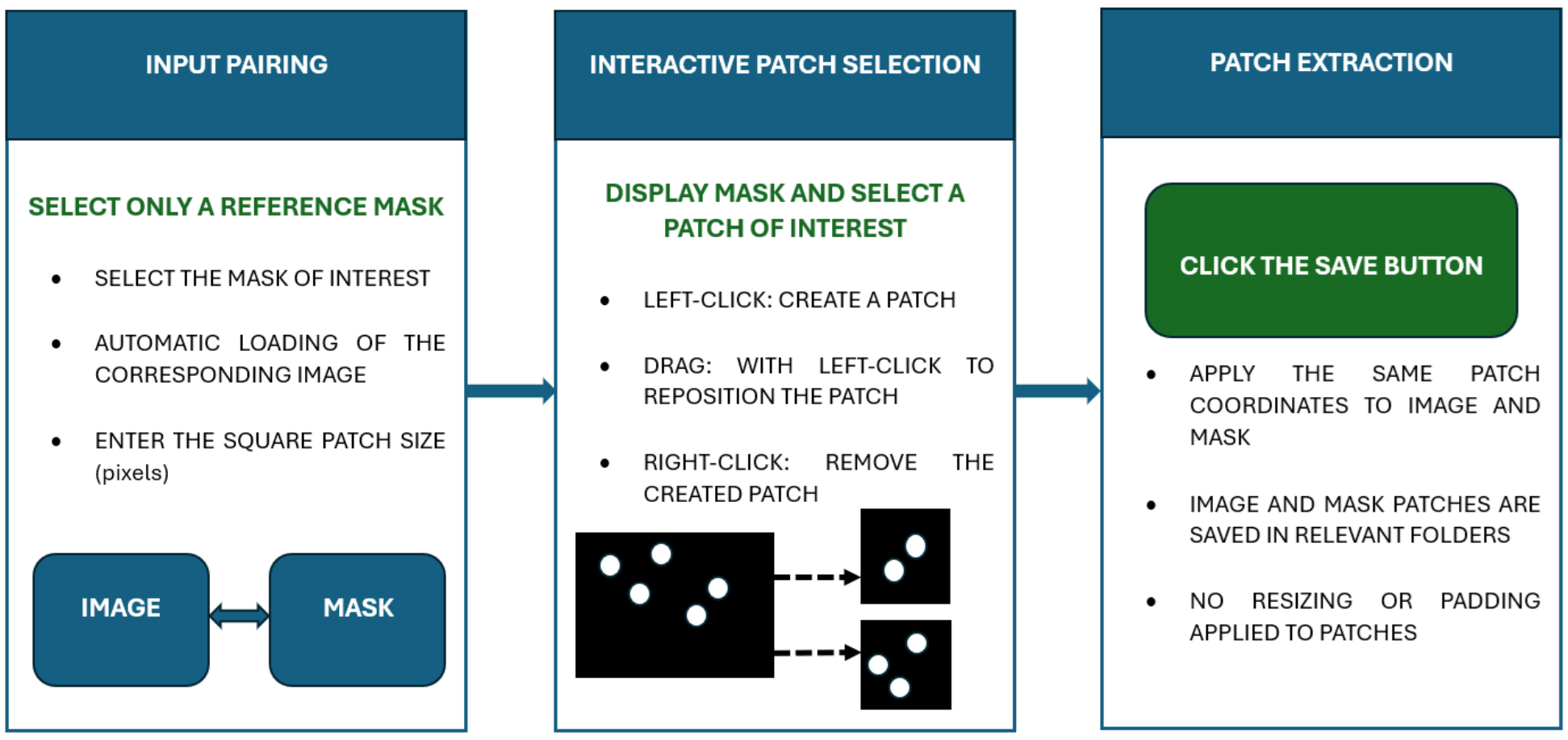

**Figure 1**: Overview of the SSP tool [54]: User selection and aligned image-mask extraction

#### 2.1.1. Selection of the patch of interest

The user selects a segmentation mask through a file-selection dialog. The script automatically loads the corresponding image from the configured image folder using the same filename and verifies that their spatial dimensions match. The user then specifies the square patch size in pixels. The mask is displayed for interactive selection. Left-clicking outside existing patches adds a patch, while left-clicking and dragging an existing patch allows its position to be adjusted. Right-clicking removes the most recently added patch. Patch positions are constrained to keep each crop entirely within the image boundaries. Selection does not depend on a predefined stride, and overlapping patches are permitted.

#### 2.1.2. Extraction of image–mask patch pairs

Clicking the Save button extracts the selected patches directly from the source image and mask using identical crop coordinates. Corresponding image and mask patches are saved as PNG files in separate output folders with matching filenames. The script also saves a text file containing the selected patch-centre coordinates, expressed as (x, y) in the source coordinate system. The extraction process performs no resizing or padding. Segmentation mask patches are cropped from the existing reference mask; the tool does not generate new foreground annotations.

### *2.2. Modified DeepLabV3+*

The DeepLabV3+ network was implemented in MATLAB R2024b [9]. Its original architecture is illustrated in [8]. Since all patch-based datasets consisted of grayscale images, the image input layer of each DeepLabV3+ configuration was replaced to accept single-channel inputs of size 640 × 640 × 1, retaining its normalization method. The first convolutional layer was also replaced to accommodate single-channel input, retaining its filter size, number of filters, padding, stride, and learning-rate factors, but using newly initialized weights. The remaining network layers were retained as generated for the selected backbone. DeepLabV3+ models with ResNet50, ResNet18, or MobileNetV2 backbones were trained separately on each of the three patch-based datasets: SSP, Overlap, and Noverlap. All models were trained on a workstation equipped with a single NVIDIA GeForce RTX 3070 GPU and an 11th-generation Intel Core i7-11700F processor operating at 2.50 GHz, with eight cores and 16 logical processors.

### 2.3. Training data

In our experiments, training data used are the same as those used in our previous work [7]. In our previous work, the image data were generated by the MiniTEM microscope by Vironova AB, Stockholm, Sweden, with an operating voltage of 25 kV and with a field of view of 3 μm for the adenovirus samples [7]. The initial data consisted of 50 MiniTEM images of size of 2048-by-2048, in a .tif format with 16 bits' depth. Annotating these images and extracting corresponding masks were done using a semi-automatic annotation software tool developed, by our previous team, at the University of Eastern Finland [7]. The resulting annotated image–mask pairs were then used to construct the SSP, Overlap, and Noverlap patch-based datasets. Figure 2 summarizes MiniTEM patch-based datasets construction and evaluation on the respective test splits and the external test set. As can be seen, the MiniTEM-based SSP dataset (no stride or stride independent) consisted of 282 image patches and their corresponding masks. Noverlap dataset (stride = patch size) consisted of 450 image patches and their corresponding masks while Overlap dataset (stride = 0.5 × patch size – *i.e., half patch size for simplification purposes*) consisted of 1250 image patches and their corresponding masks. All patches had the same size of 640 x 640 pixels.

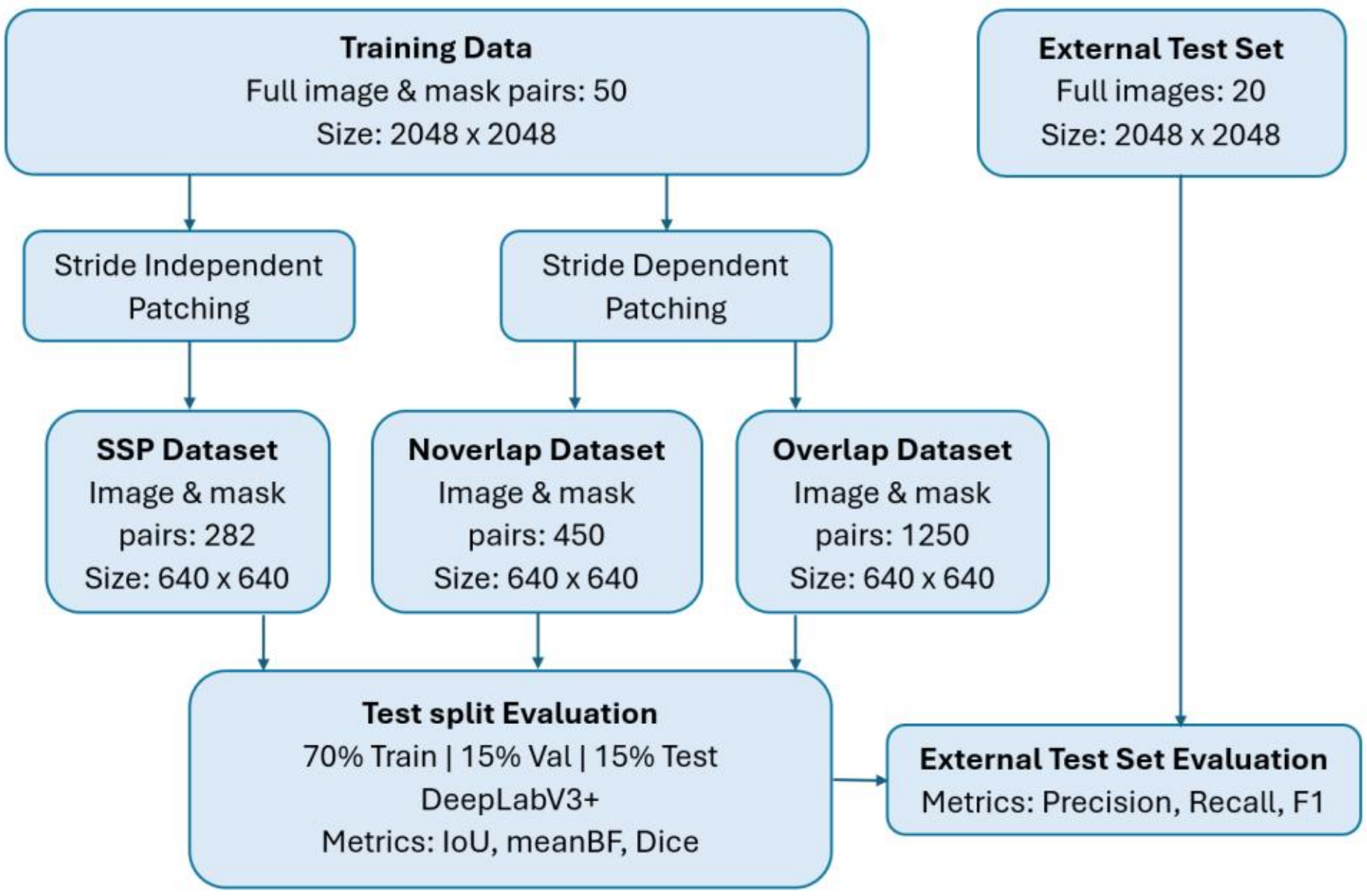

**Figure 2**: Overview of MiniTEM patch-based dataset construction and evaluation

Table 1 presents the number of patches and the unique image area covered or omitted by each patching strategy. Each patch measures 640 × 640 pixels. Noverlap and Overlap values assume extraction from the top-left corner without padding or border adjustment. SSP coverage varies with user-selected patch locations.

**Table 1**: Patch counts and spatial coverage per 2048 × 2048 source image for the three patching strategies

| Technique | Stride | Patches per image | Unique Area covered | Area omitted |
|---|---|---|---|---|
| SSP | Not Applicable | Variable | Variable | Variable |
| Noverlap | 640 pixels | 9 | 1920 × 1920 pixels | 507,904 pixels (12.11%) |
| Overlap | 320 pixels | 25 | 1920 × 1920 pixels | 507,904 pixels (12.11%) |

After patching, in each dataset case, the training set split was 70% of the training dataset, the validation set split was 15% of the training dataset, and the test set split was 15% of the training dataset.

### *2.4. External test set*

A dataset containing 20 full images each measuring 2048 × 2048 pixels was used as an external test set to evaluate the performance of trained models [7]. The trained models were applied directly to these full-resolution images. DeepLabV3+ models with ResNet50, ResNet18, and MobileNetV2 backbones were assessed for their ability to segment intact adenoviruses on the external test set. Visual assessment of correctly identified intact adenoviruses were counted as true positives (TP), unmatched predicted objects as false positives (FP), and missed intact adenoviruses as false negatives (FN). Precision (Eq. 1), recall (Eq. 2), and the F1-score (Eq. 3) [13] were calculated from these counts.

$$precision = \frac{TP}{TP+FP} \quad (1)$$

$$recall = \frac{TP}{TP+FN} \quad (2)$$

$$f_{value} = \frac{2*recall*precision}{recall+precision} \quad (3)$$

### *2.5. Hyperparameter settings*

Training hyperparameters were manually adjusted with reference to previous studies [10–12]. No further adjustments were made if training accuracy reached 90% within the first 10% of the total epochs. Except for the image input layer modification described in Section 2.2, the architecture settings were retained. DeepLabV3+ models were trained for 30 epochs using a mini-batch size of 2, an initial learning rate of 0.0001, an L2 regularization factor of 0.00005, and the Adam optimizer (adaptive moment estimation) [55].

### *2.6. Augmentation operations*

Data augmentation was applied to increase training-data variability and reduce overfitting. The operations included random rotation between −45° and 45°, uniform scaling by a factor between 0.8 and 1.2, horizontal and vertical translation between −10 and 10 pixels, and horizontal and vertical shearing between −15° and 15° [11–13]. These geometric transformations were applied to each image and its corresponding mask. Brightness adjustment was applied only to the image by multiplying its intensities by a randomly selected factor from {0.90, 0.95, 1.00, 1.05, 1.10} [11–13]. Geometric transformations generally require interpolation to estimate pixel values at transformed coordinates, with the choice of interpolation method influencing image intensities and mask label assignments [11], [33–42]. Previous studies investigated grayscale image interpolation methods relevant to geometric transformations [14], [43], [44]. Other studies examined the influence of rounding functions on interpolation accuracy, demonstrating that performance can be significantly affected by the rounding strategy [15–17], which can also affect the quality of transformed image. A conceptual framework also classified interpolation algorithms into extra-pixel and non-extra-pixel methods [18]. Further research explored adaptive optimization, weighting schemes, and extrapolation techniques to improve interpolation algorithms [19–24], [43], [45-49]. Optimized efforts for interpolation algorithms were also extended to practical cardiac imaging software [17], [25]. Together, these studies highlight the importance of interpolation in image processing and its relevance to data augmentation for deep learning. Although previous studies reported benefits of extra-pixel interpolation over nearest-neighbour interpolation [11], [13], this demonstration used nearest-neighbour interpolation for both images and masks during geometric augmentation to limit computational overhead during training on a single GPU.

## 3. Results

Experimental results are presented in two subsections covering quantitative evaluations on the test split and the external test set.

### *3.1. Quantitative evaluation on the test split*

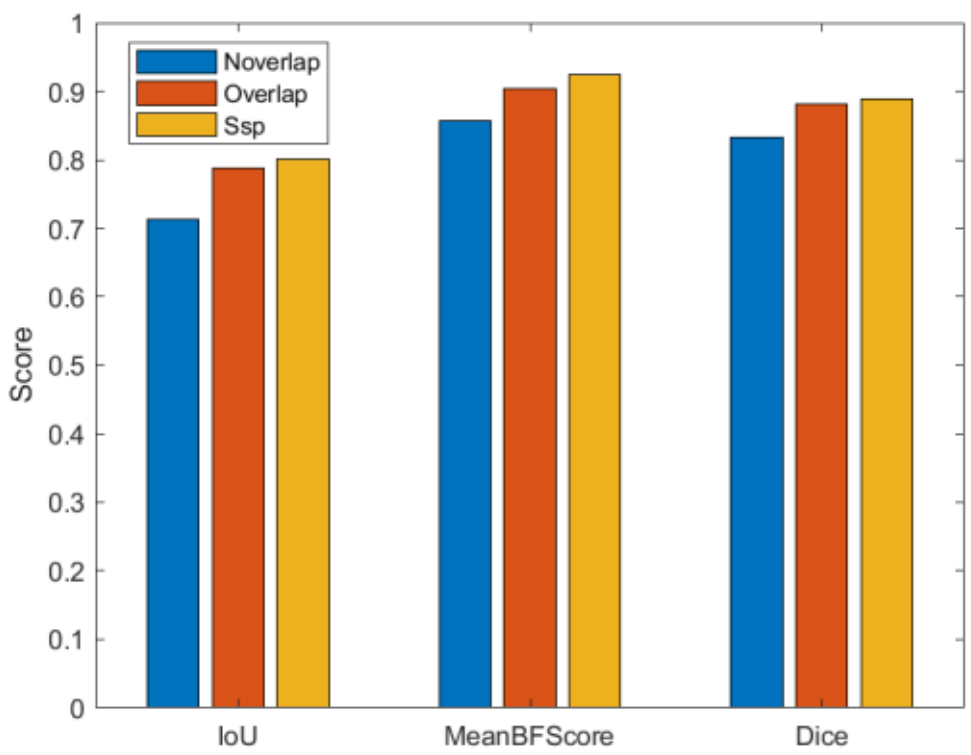

**Figure 3.** DeepLabV3+ with ResNet50

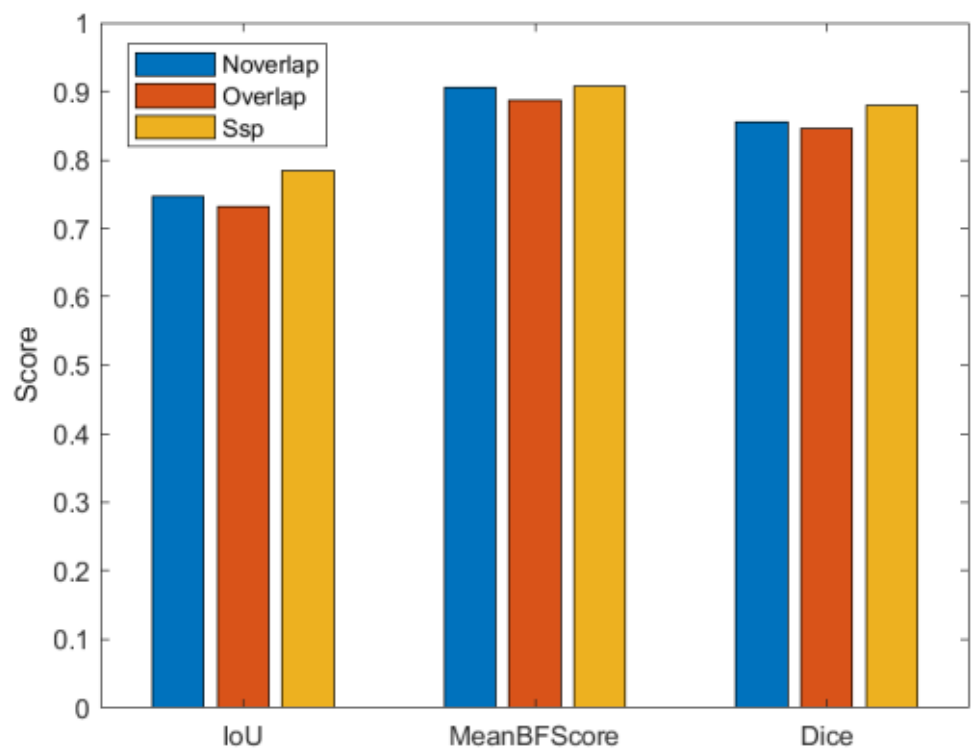

**Figure 4**. DeepLabV3+ with ResNet18

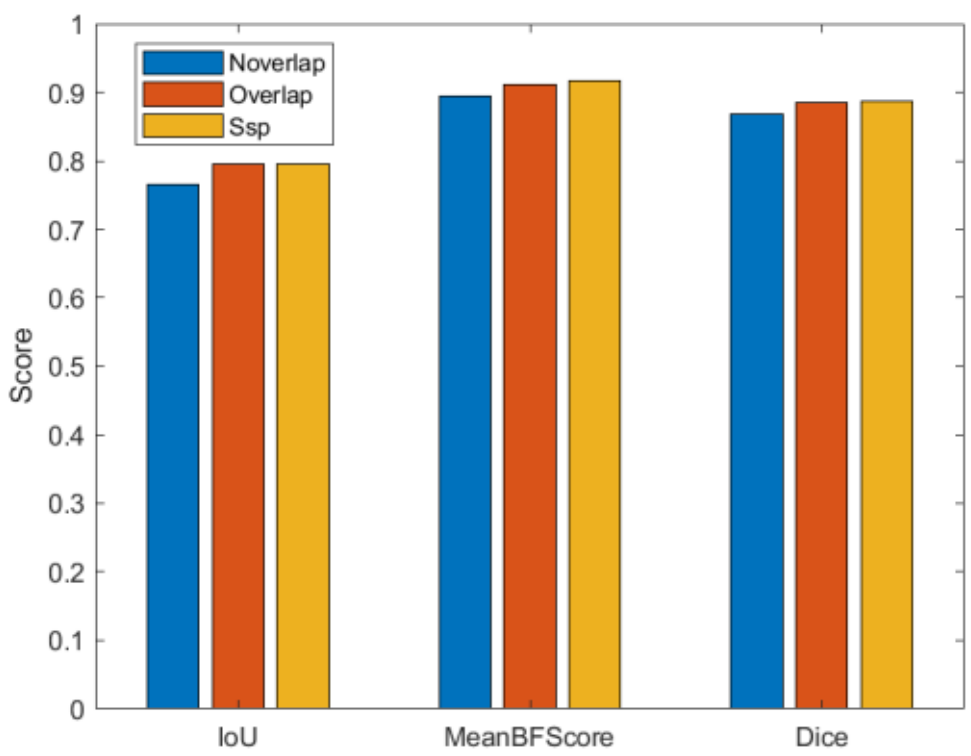

**Figure 5.** DeepLabV3+ with MobileNetV2

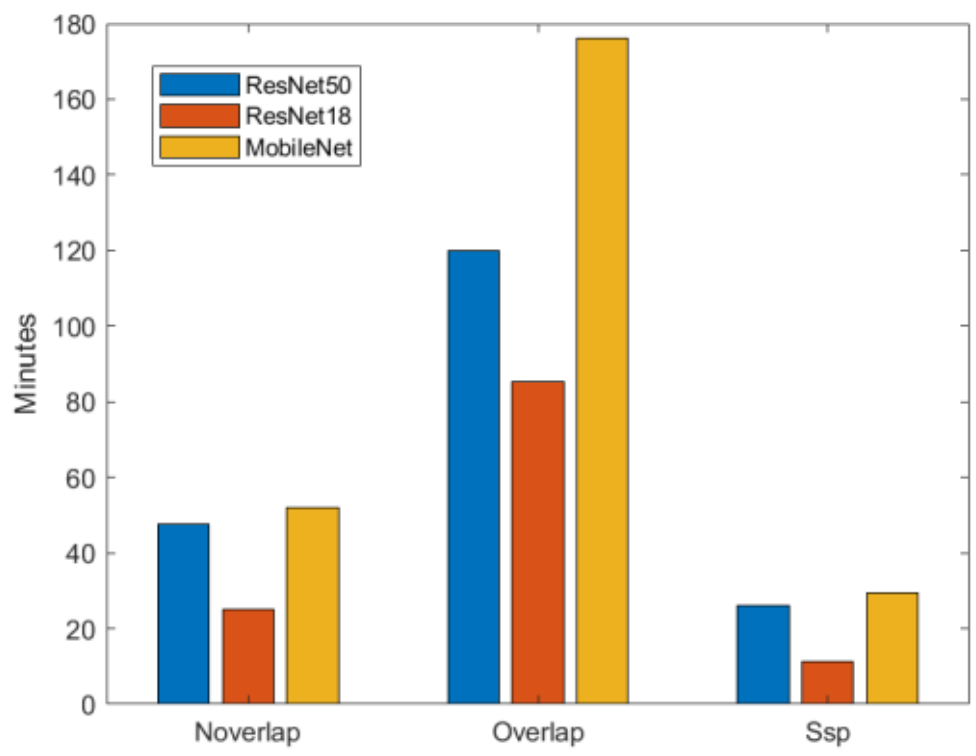

**Figure 6**. DeepLabV3+ training time

### *3.2. Quantitative evaluation on the external test set*

**Table 2**. Quantitative evaluation of DeepLabV3+ models with ResNet50, ResNet18, and MobileNetV2 backbones on the external test set. Counts shown as n / 20 denote the total number of predicted objects across the 20 external test images.

| | TP | FP | FN | Precision | Recall | F-value | Av. Prec. | Av. Rec | Av. F-v |
|---|---|---|---|---|---|---|---|---|---|
| Noverlap_R50: 179 / 20 | 153 | 26 | 0 | 0.8547 | 1.0000 | 0.9217 | 0.9470 | **0.9932** | 0.9682 |
| Noverlap_R18: 146 / 20 | 144 | 2 | 2 | 0.9863 | 0.9863 | 0.9863 | | | |
| Noverlap_MOB: 145/ 20 | 145 | 0 | 1 | 1.0000 | 0.9932 | 0.9966 | | | |
| Overlap_R50: 143 / 20 | 141 | 2 | 3 | 0.9860 | 0.9792 | 0.9826 | 0.9597 | 0.9838 | 0.9711 |
| Overlap_R18: 141 / 20 | 139 | 2 | 3 | 0.9858 | 0.9789 | 0.9823 | | | |
| Overlap_MOB: 162 / 20 | 147 | 15 | 1 | 0.9074 | 0.9932 | 0.9484 | | | |
| Ssp_R50: 156 / 20 | 151 | 5 | 0 | 0.9679 | 1.0000 | 0.9837 | **0.9731** | 0.9860 | **0.9795** |
| Ssp_R18: 145 / 20 | 142 | 3 | 2 | 0.9793 | 0.9861 | 0.9827 | | | |
| Ssp_MOB: 143 / 20 | 139 | 4 | 4 | 0.9720 | 0.9720 | 0.9720 | | | |

## 4. Discussion

Figures 3–5 show that SSP generally achieved the highest segmentation scores across the three backbones, although its IoU and Dice scores were similar to those of Overlap for MobileNetV2. Figure 6 shows that SSP required the shortest training time for every backbone. Together, these results suggest that SSP offers a favorable balance between segmentation performance and training time under the evaluated conditions. Its shorter training time should also be considered alongside its smaller training dataset: 282 patches, compared with 450 for Noverlap and 1250 for Overlap.

Table 1 further illustrates the spatial coverage of the patching strategies. Noverlap and Overlap generated 9 and 25 patches per source image, respectively, but both covered the same 1920 × 1920 region, leaving 12.11% of each image outside the extracted patches. Thus, overlapping patches increased the number of training samples without extending coverage to the remaining borders. SSP instead allowed users to select patches around objects of interest, with variable spatial coverage. Considered alongside the segmentation results, these differences support the value of object-focused patch selection rather than simply generating more patches.

Table 2 shows that SSP achieved the highest average precision (0.9731) and F1-score (0.9795) across the three backbones, suggesting a favorable balance between correctly detecting intact adenoviruses and limiting false positives. Noverlap achieved the highest average recall (0.9932), but its lower precision (0.9470) indicates a greater tendency to include false positives. Overlap achieved higher average precision and F1-score than Noverlap, although its average recall was lower. Overall, these averages support SSP as the strongest patching strategy in terms of precision and F1-score under the evaluated conditions, while Noverlap favored recall.

## 5. Conclusions

SSP's user-guided selection of patches around objects of interest can provide a favorable balance between segmentation performance and model training time. SSP generally achieved higher segmentation scores on the test splits and the shortest training time across all three backbones. Its highest average precision and F1-score on the external test set further support its potential, although Noverlap retained an advantage in average recall. Although Noverlap and Overlap omitted only 12.11% of each source image, potentially less than SSP, this did not generally translate into better segmentation performance. These results establish stride-independent patching as a promising alternative to stride-dependent patching for the evaluated task. Its practical value must nevertheless be weighed against the user effort required for patch selection. Future work should assess that effort and performance across broader datasets.

**Supplementary Materials:** The SSP tool is available at https://doi.org/10.5281/zenodo.23403499

**Author Contributions:** Conceptualization, O.R.; methodology, O.R.; software, O.R.; validation, O.R.; formal analysis, O.R.; investigation, O.R.; resources, O.R.; data curation, O.R.; writing—original draft preparation, O.R.; writing—review and editing, O.R.; visualization, O.R.; supervision, O.R.; project administration, O.R. The author has read and agreed to the published version of the manuscript.

**Funding:** This research received no external funding.

**Data Availability Statement:** The data supporting this study are not publicly available but may be provided upon request, subject to approval by the University of Eastern Finland and FinVector.

**Acknowledgments:** The author thanks researchers at the University of Eastern Finland and FinVector for providing the training data used in the experiments, and the University of Limerick for providing additional resources.

**Conflicts of Interest:** The author declares no conflicts of interest.

## References

1. C. Yang, X. Lu, Z. Lin, E. Shechtman, O. Wang and H. Li, "High-Resolution Image Inpainting Using Multi-scale Neural Patch Synthesis," 2017 IEEE Conference on Computer Vision and Pattern Recognition (CVPR), Honolulu, HI, USA, 2017, pp. 4076-4084, doi: 10.1109/CVPR.2017.434
2. Benjamin Bergner, Christoph Lippert, Aravindh Mahendran, Iterative Patch Selection for High-Resolution Image Recognition, arXiv:2210.13007 [cs.CV]
3. Pielawski N, Wählby C (2020) Introducing Hann windows for reducing edge-effects in patch-based image segmentation. PLoS ONE 15(3): e0229839
4. E. Zhang et al., Adaptive Patching for High-resolution Image Segmentation with Transformers, International Conference for High Performance Computing, Networking, Storage and Analysis, Atlanta, GA, USA, 2024, pp. 1-16
5. Kesidis, A.L.; Krassanakis, V.; Misthos, L.-M.; Merlemis, N. patchIT: A Multipurpose Patch Creation Tool for Image Processing Applications. Multimodal Technol. Interact. 2022, 6, 111
6. DeepAI, "Stride," DeepAI, Accessed: Mar. 20, 2025. [Online]. Available: https://deepai.org/machine-learning-glossary-and-terms/stride
7. Rukundo, O., Behanova, A., De Feo, R. et al. Convolutional Neural Networks for Automatic Detection of Intact Adenovirus from TEM Imaging with Debris, Broken and Artefacts Particles. *ArXiv*, 2023: arXiv preprint arXiv: 2310.19630, pp. 1–13.
8. Chen, L., Y. Zhu, G. Papandreou, F. Schroff, and H. Adam. "Encoder-Decoder with Atrous Separable Convolution for Semantic Image Segmentation." Computer Vision — ECCV 2018, 833-851. Munic, Germany: ECCV, 2018.
9. MathWorks. 2024. "DeepLab v3+ Layers." MATLAB Documentation. https://uk.mathworks.com/help/releases/R2024b/vision/ref/deeplabv3pluslayers.html
10. Rukundo, O. Effects of Image Size on Deep Learning. Electronics 2023, 12, 985
11. Rukundo, O. Evaluation of extra pixel interpolation with mask processing for medical image segmentation with deep learning. SIViP 18, 7703–7710 (2024)
12. Rukundo, O. Effect of the regularization hyperparameter on deep learning-based segmentation in LGE-MRI. In Proceedings of the SPIE/COS Photonics Asia, Nantong, China, 10–20 October 2021; Volume 11897
13. Rukundo, O., Beyond Nearest Neighbor Interpolation in Data Augmentation, SIViP 20, 525 (2026)
14. Rukundo, O. Optimal Methods Research on Grayscale Image Interpolation; China National Knowledge Infrastructure CNKI, TP391.41: Beijing, China, 2012
15. Rukundo, O. Effects of improved-floor function on the accuracy of bilinear interpolation algorithm. Comput. Inf. Sci. 2015, 8
16. Rukundo, O. Evaluation of Rounding Functions in Nearest Neighbour Interpolation. Int. J. Comput. Methods 2021, 18, 2150024
17. Rukundo, O.; Schmidt, S. Stochastic Rounding for Image Interpolation and Scan Conversion. Int. J. Adv. Comput. Sci. Appl. 2022, 13, 13–22
18. Rukundo, O. Non-extra pixel interpolation. Int. J. Image Graph. 2020, 20, 2050031
19. Rukundo, O.; Huang, M.; Cao, H. Optimization of Bilinear Interpolation Based on Ant Colony Algorithm. In Proceedings of the 2nd International Conference Electrical and Electronics Engineering, Macao, China, 1–2 December 2011; pp. 571–580
20. Rukundo, O.; Wu, K.; Cao, H. Image Interpolation Based on The Pixel Value Corresponding to The Smallest Absolute Difference. In Proceedings of the 4th International Workshop on Advanced Computational Intelligence, Wuhan, China, 19–21 October 2011; pp. 434–437
21. Rukundo, O.; Maharaj, B. Optimization of Image Interpolation based on Nearest Neighbour Algorithm. In Proceedings of the 9th International Conference on Computer Vision Theory and Applications, Lisbon, Portugal, 5–8 January 2014; pp. 641–647
22. Rukundo, O. Normalized weighting schemes for image interpolation algorithms. Appl. Sci. 2023, 13, 1741
23. Rukundo, O.; Cao, H. Advances on image interpolation based on ant colony algorithm. SpringerPlus 2016, 5, 403
24. Rukundo, O.; Schmidt, S. Extrapolation for image interpolation. In Proceedings of the SPIE/COS Photonics Asia, Beijing, China, 11–13 October 2018; Volume 10817
25. Rukundo, O.; Schmidt, S.E.; Von Ramm, O.T. Software implementation of optimized bicubic interpolated scan conversion in echocardiography. arXiv 2020, arXiv:2005.11269
26. D. K. Gupta, G. Mago, A. Chavan, and D. K. Prasad, "Patch Gradient Descent: Training Neural Networks on Very Large Images," arXiv preprint arXiv:2301.13817, 2023
27. M. van Rijthoven, M. Balkenhol, K. Siliņa, J. van der Laak, and F. Ciompi, "HookNet: Multi-resolution convolutional neural networks for semantic segmentation in histopathology whole-slide images," Medical Image Analysis, 68 (101890), 2021
28. R. J. Chen, C. Chen, Y. Li, T. Y. Chen, A. D. Trister, R. G. Krishnan, and F. Mahmood, "Scaling Vision Transformers to Gigapixel Images via Hierarchical Self-Supervised Learning," in Proceedings of the IEEE/CVF Conference on Computer Vision and Pattern Recognition (CVPR), 2022, pp. 16144–16155
29. H. Pinckaers, B. van Ginneken, and G. Litjens, "Streaming Convolutional Neural Networks for End-to-End Learning With Multi-Megapixel Images," IEEE Transactions on Pattern Analysis and Machine Intelligence, vol. 44, no. 3, pp. 1581–1590, 2022
30. N. Gourmelon, T. Seehaus, M. Braun, A. Maier, and V. Christlein, "Calving fronts and where to find them: a benchmark dataset and methodology for automatic glacier calving front extraction from synthetic aperture radar imagery," Earth System Science Data, vol. 14, pp. 4287–4313, 2022
31. S. M. Azimi, D. Britz, M. Engstler, M. Fritz, and F. Mücklich, "Advanced Steel Microstructural Classification by Deep Learning Methods," Scientific Reports, vol. 8, Art. no. 2128, 2018

32. Damian J. Matuszewski, Ida-Maria Sintorn, TEM virus images: Benchmark dataset and deep learning classification, Computer Methods and Programs in Biomedicine, Volume 209, 2021, 106318, ISSN 0169-2607
33. Rukundo, O., "Toward Optimal Adenovirus Detection Using YOLO26," arXiv preprint arXiv:2607.17799, 2026
34. P. Thévenaz, T. Blu, and M. Unser, "Interpolation revisited," IEEE Transactions on Medical Imaging, 19 (7), pp. 739–758, 2000
35. B. Zitová and J. Flusser, "Image registration methods: a survey," Image and Vision Computing, 21( 11), pp. 977–1000, 2003
36. W. Kim, Y. Kang, S. Lee, H.-J. Lee, and Y. Nam, "Label-Preserving Data Augmentation for Robust Segmentation of Thin Structure in MRI," Investigative Magnetic Resonance Imaging, vol. 28, no. 3, pp. 107–113, 2024
37. B. Murray et al., "Lazy Resampling: Fast and information preserving preprocessing for deep learning," Computer Methods and Programs in Biomedicine, vol. 257, Art. no. 108422, 2024
38. L. Henschel, D. Kügler, and M. Reuter, "FastSurferVINN: Building resolution-independence into deep learning segmentation methods—A solution for HighRes brain MRI," NeuroImage, vol. 251, Art. no. 118933, 2022
39. Lehmann, T. M.; Gönner, C.; Spitzer, K. Survey: Interpolation methods in medical image processing. IEEE Transactions on Medical Imaging 18(11), 1049–1075 (1999)
40. Keys, R. G. Cubic convolution interpolation for digital image processing. IEEE Transactions on Acoustics, Speech, and Signal Processing 29(6), 1153–1160 (1981)
41. Lehmann, T. M.; Gönner, C.; Spitzer, K. Addendum: B-spline interpolation in medical image processing. IEEE Transactions on Medical Imaging 20(7), 660–665 (2001)
42. Blu, T.; Thévenaz, P.; Unser, M., "Linear interpolation revitalized," IEEE Transactions on Image Processing 13(5), 673–681 (2004)
43. Li, X.; Orchard, M. T. New edge-directed interpolation. IEEE Transactions on Image Processing 10(10), 1521–1527 (2001)
44. Grevera, G. J.; Udupa, J. K. An objective comparison of 3-D image interpolation methods. IEEE Transactions on Medical Imaging 17(4), 642–652 (1998)
45. Zhang, L.; Wu, X. An edge-guided image interpolation algorithm via directional filtering and data fusion. IEEE Transactions on Image Processing 15(8), 2226–2238 (2006)
46. Giachetti, A.; Asuni, N. Real-time artifact-free image upscaling. IEEE Transactions on Image Processing 20(10), 2760–2768 (2011)
47. Romano, Y.; Protter, M.; Elad, M. Single image interpolation via adaptive nonlocal sparsity-based modeling. IEEE Transactions on Image Processing 23(7), 3085–3098 (2014)
48. Zheng, J.; Song, W.; Wu, Y.; Liu, F. Image interpolation with adaptive k-nearest neighbours search and random non-linear regression. IET Image Processing 14(8), 1539–1548 (2020)
49. Zhang, J. Weighted-encoding-based image interpolation with the nonlocal linear regression model. Applied Optics 59(28), 8588–8594 (2020)
50. Orhan, S.; Baştanlar, Y. Training CNNs with image patches for object localisation. Electronics Letters 54(7), 424–426 (2018)
51. Agarwal, R.; Diaz, O.; Lladó, X.; Yap, M. H.; Martí, R. Automatic mass detection in mammograms using deep convolutional neural networks. Journal of Medical Imaging 6(3), 031409 (2019)
52. De Carvalho, O. L. F., et al. Instance Segmentation for Large, Multi-Channel Remote Sensing Imagery Using Mask-RCNN and a Mosaicking Approach. Remote Sensing 13(1), 39 (2021)
53. Roszkowiak, L.; Korzynska, A.; Siemion, K.; et al. System for quantitative evaluation of DAB&H-stained breast cancer biopsy digital images (CHISEL). Scientific Reports 11, 9291 (2021)
54. Olivier Rukundo, Ph.D. (2026). Semi-automatic Stride-independent Patching (Version v1.0.0) [Computer software]. Zenodo. https://doi.org/10.5281/zenodo.23403499
55. D. P. Kingma and J. Ba, "Adam: A Method for Stochastic Optimization," arXiv preprint arXiv:1412.6980, 2014. https://doi.org/10.48550/arXiv.1412.6980